\documentclass[letterpaper, 10 pt, conference]{packages/ieeeconf}

\IEEEoverridecommandlockouts                              

\usepackage{times}
\usepackage{epsfig}
\usepackage{graphicx}
\usepackage{amsmath}
\usepackage{amssymb}
\usepackage{subfigure} 
\usepackage[ruled]{algorithm}
\usepackage{algpseudocode}
\usepackage{ctable}
\usepackage[export]{adjustbox}
\usepackage{import}
\usepackage{pgf}

\makeatletter
\let\NAT@parse\undefined
\makeatother
\usepackage[numbers]{natbib}

\usepackage{packages/sparo_acronyms}
\usepackage{packages/sparo_math}
\usepackage{packages/sparo_SIunits}
\usepackage{packages/sparo_misc}
\usepackage{multirow}

\usepackage{soul,color}
\usepackage{verbatim} 

\usepackage{xcolor}

\DeclareMathOperator*{\argmin}{argmin}

\usepackage[symbol]{footmisc}

\title{\LARGE \bf
	Robust Imaging Sonar-based Place Recognition and  Localization \\ in Underwater Environments
}

\author{Hogyun Kim$^{1}$ Gilhwan Kang$^{1}$ Seokhwan Jeong$^{1}$ Seungjun Ma$^{1}$ and Younggun Cho$^{1*}$
	\thanks{This work is supported by Inha University, Institute of Information \& communications Technology Planning \& Evaluation (IITP) grant funded by the Korea government(MSIT) (No.2022-0-00448, Deep Total Recall), and the Korea Institute of Marine Science and Technology Promotion (KIMST), funded by the Ministry of Oceans and Fisheries (20210562). Also, this was supported by the National Research Foundation of Korea(NRF) grant funded by the Korea government(MSIT) (No.2021R1G1A1009941).}
	\thanks{$^{1}$H. Kim, $^{1}$G. Kang, $^{1}$S. Jeong, $^{1}$S. Ma, and $^{1*}$Y. Cho are with Dept. Electrical and Computer Engineering, Inha University, Incheon, South Korea
		{\tt\small [12170550, 22222151, 12171433, richard7714]@inha.edu, yg.cho@inha.ac.kr}}%
}

\begin{document}

\maketitle
\thispagestyle{empty}
\pagestyle{empty}

\begin{abstract}
%

Place recognition using \ac{SONAR} images is an important task for \ac{SLAM} in underwater environments. This paper proposes a robust and efficient imaging SONAR-based place recognition, SONAR context, and loop closure method. Unlike previous methods, our approach encodes geometric information based on the characteristics of raw SONAR measurements without prior knowledge or training. We also design a hierarchical searching procedure for fast retrieval of candidate SONAR frames and apply adaptive shifting and padding to achieve robust matching on rotation and translation changes. In addition, we can derive the initial pose through adaptive shifting and apply it to the \ac{ICP}-based loop closure factor. We evaluate the SONAR context's performance in the various underwater sequences such as simulated open water, real water tank, and real underwater environments. The proposed approach shows the robustness and improvements of place recognition on various datasets and evaluation metrics. Supplementary materials are available at \verb|https://github.com/sparolab/sonar_context.git|.

\end{abstract}

\section{introduction}

Robust place recognition is essential for long-term operation and accurate state estimation of an \ac{AUV}. Especially in \ac{SLAM} problems,  precise loop closure critically affects the overall performance and quality of robot states and global maps. For aerial and ground robotics, there are several well-known methods with vision \cite{cummins2008fab, galvez2012bags} and \ac{LiDAR}-based \cite{uy2018pointnetvlad, dube2018segmap, kim2018sc} sensors. Additionally, \ac{GPS} information can be a powerful alternative or prior measurement for most of the above methods.

 However, in underwater environments, place recognition using optical sensors presents several challenges due to such environments' distinctive attributes. For instance, water turbidity can disturb optical sensors and electromagnetic wave attenuation can hinder \ac{GPS} utilization.

To overcome these limitations, Imaging \ac{SONAR} is one of the majorly used perceptual sensors for the navigation of an \ac{AUV}. Unlike optical sensors, \ac{SONAR} employs the reflection of acoustic waves to generate a \ac{SONAR} image. Because sound spreads farther than light in underwater environments, an extensive sensing range can be acquired. However, \ac{SONAR} also has drawbacks, such as elevation loss, observational uncertainty, and a low signal-to-noise ratio \cite{wang2021robust}. 
Because these characteristics discourage applying a traditional place recognition methodology, a tailormade approach for the underwater environment is needed.

\begin{figure}[t!]
	\centering
	\def\width{0.95\columnwidth}%
	\includegraphics[clip, trim= 85 25 85 10, width=\width]{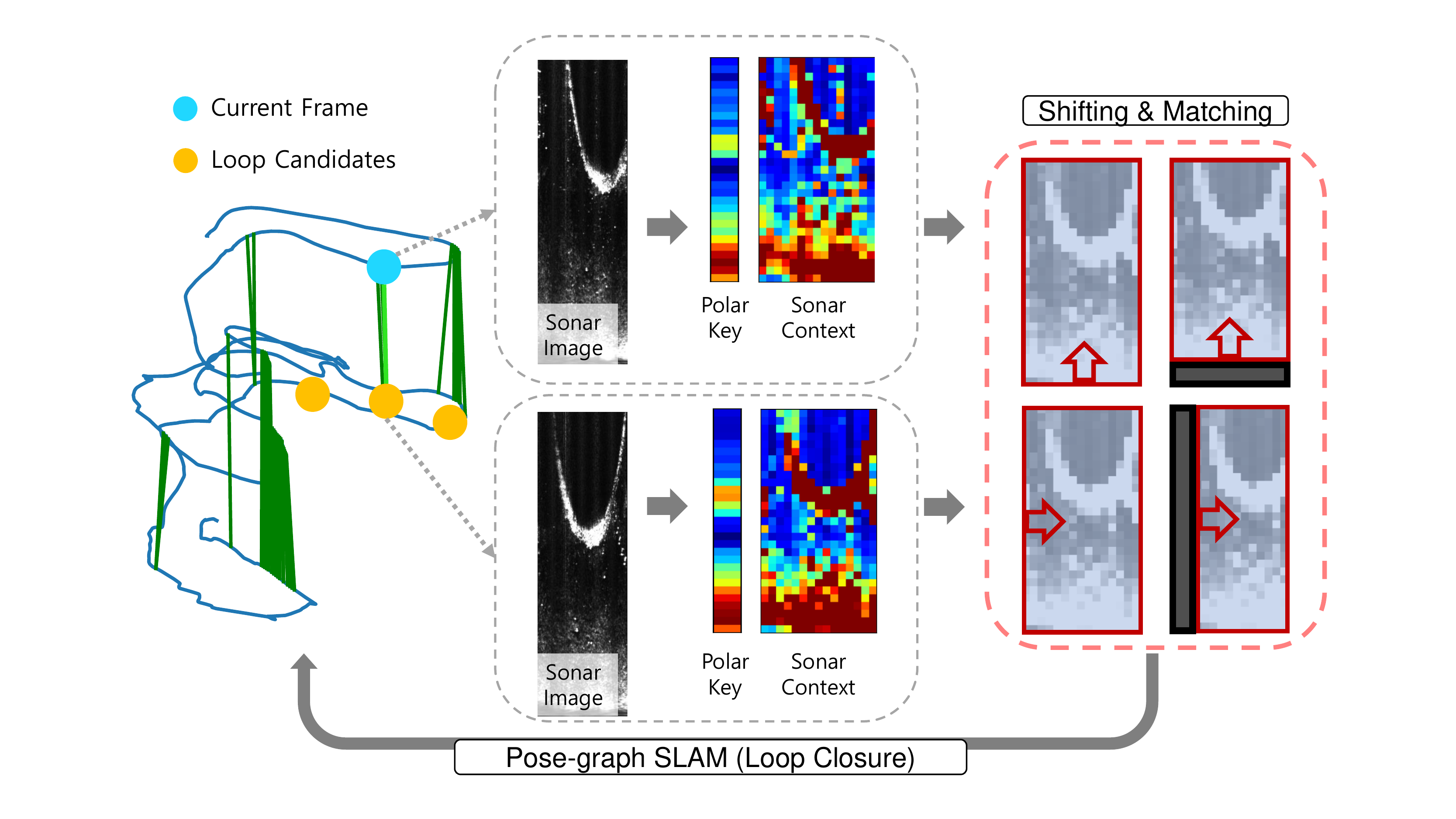}
	\caption{Our place recognition method using \ac{SONAR} context, polar key and adaptive shifting. The figure on the left shows qualitative evaluation including trajectory with current frame and loop candidates in the ARACATI 2017 dataset \cite{dos2022cross}. Candidates are selected through polar key, and adaptive shifting is applied between \ac{SONAR} contexts to match. More details of our method are shown in \figref{fig:our_slam_figure}.
    }
	\label{fig:main}
    \vspace{-0.35cm}
\end{figure}

Many researchers have tried to utilize \ac{SONAR} in place recognition based on the classic optical feature method \cite{westman2018feature,li2018pose,oliveira2021performance,li2022acoustic}. These approaches are suitable for retrieving local correspondences. However, global localization with optical features often suffers from low precision of loop detection because of insufficient geometric and structural information in underwater environments. Nevertheless, many feature-based methods still utilize loop closure detection with nearest neighbor search by means of the Euclidean distance of robot poses or specific approaches that rely on environmental constraints and assumptions. Hence, precise loop detection is essential for the reliable operation of an \ac{AUV} in unknown underwater environments. 

In this paper, we propose a novel and precise Imaging \ac{SONAR}-based place recognition as shown in \figref{fig:main}. We design a global descriptor that encodes geometric and intensity characteristics for loop closure. Focusing on the characteristics of SONAR measurements, our method utilizes SONAR measurements in polar coordinates and embeds descriptors without heavy computation. To improve the performance of place retrieval, we propose adaptive feature shifting and matching algorithms. 

Our main contribution points in this paper can be summarized as follows.
\begin{enumerate}
    \item We propose a precise \ac{SONAR}-based global descriptor, that can encode geometric characteristics of underwater environments. The descriptor consists of coarse (polar key) and fine description (SONAR Context) for efficient loop closure detection. 
    \item By considering the nature of SONAR measurements, we develop a robust descriptor for rotational and translational differences through adaptive shifting and matching algorithms.
    \item The proposed method estimates the initial pose for \ac{ICP}, which leads to better loop-closing performance.
    \item We show our comprehensive experiments in simulation, real water tank, and real ocean environments with different structural characteristics.
\end{enumerate}

We arrange the rest of our paper as follows: Section II describes related works. Section III depicts the detailed methodology of \ac{SONAR}-based place recognition and loop closure. Section IV consists of various evaluations of our methods. Finally, the conclusion in Section V consists of the summary and future works. 
\section{related works}

Many researchers have studied SONAR-based SLAM for decades, and there are two main approaches: a local descriptor-based \ac{SLAM} that uses each frame's feature, and a global descriptor-based \ac{SLAM} which uses a representative of each frame. This section focuses on SONAR-based place recognition in the above research.

The traditional method \cite{oliveira2021performance} extracts features, creates local descriptors, and recognizes revisited places through the nearest-neighbor search algorithm. However, this method is highly vulnerable in SONAR images without features. To make up for this shortcoming, \citet{tang2020sonar} proposed an image mosaic method, which simply increases the number of features. Much like the aforementioned method, this one is only useful in an environment with abundant features.

\citet{lee2014experimental} leveraged various artificial landmarks, allowing themselves to implement their \ac{SLAM} algorithms. 
In addition, \citet{xu2022robust} conducted \ac{SLAM} using a Jacobian matrix generated from local descriptors based on features or landmarks.
These methods are also difficult to utilize in underwater environments where prior knowledge of landmarks, environments, and abundant features cannot be assumed.

Recently, a learning-based SONAR-based \ac{SLAM} has emerged. \citet{li2018pose} applied learning-based saliency detection as a global descriptor to make their method robust in underwater environments and conduct pose-graph \ac{SLAM}. 
Furthermore, \citet{ribeiro2018underwater} proposed a place recognition method that uses feature extraction based on convolutional neural networks and matching based on a Triplet Distance-Based Logistic Network (Triplet-DBL-Net). These methods can be utilized as a global descriptor, but they cannot be used in real time due to memory problems and computational speed. 

To enable real-time use of global descriptors, one SONAR-based \ac{SLAM} is complemented by other modalities. 
The most representative SONAR-based \ac{SLAM} to compensate for SONAR's shortcomings is an opti-acoustic-based one that leverages an optical camera. With an optical camera, a visual-based global descriptor such as \cite{jang2021multi} can be used for the SONAR-based \ac{SLAM} method. 
However, finding matching pairs in large-scale environments is difficult because of the extreme range disparity between SONAR and camera images.

Moreover, \citet{dos2022cross} suggested cross-view and cross-domain underwater place recognition. An \ac{AUV} can conduct \ac{SLAM} employing a particle filter and recognize the revisited place by matching acoustic images with segmented aerial georeferenced images a drone or satellite has acquired. That is, in this method, the aerial device (other modality) is also essential for generating a global descriptor.

\section{background}

\subsection{sonar representation}

\begin{figure}[h]
    \centering
    \def\width{\columnwidth}%
    \subfigure[Original polar image]
    {
	\includegraphics[clip, trim=130 20 100 20, width=0.45\width]{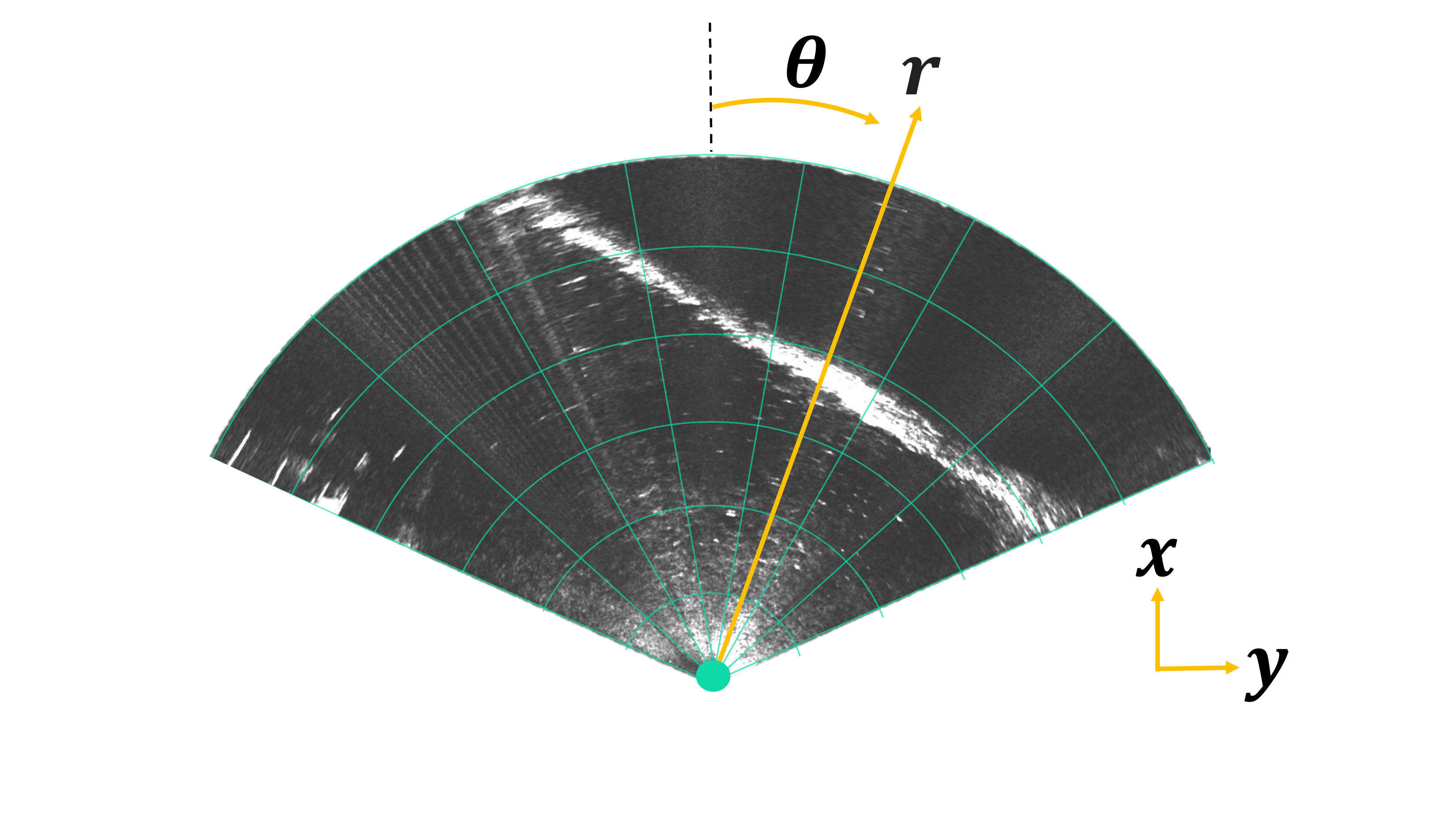}
    \label{fig:SONAR_geometry:polar}
    }
    \subfigure[Encoded polar image]{
	\includegraphics[clip, trim=150 10 45 10, width=0.45\width]{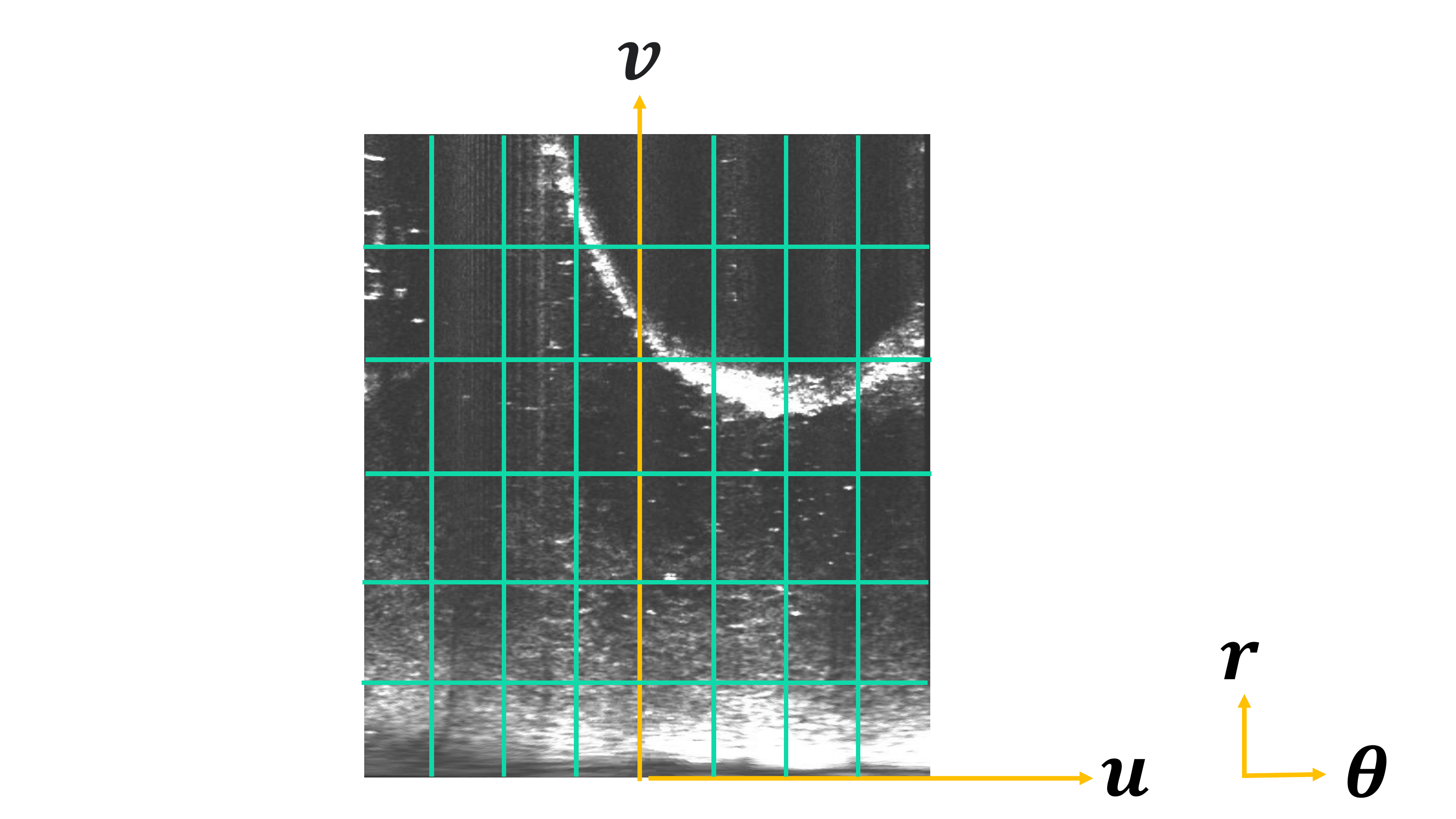}
    \label{fig:SONAR_geometry:encode_polar}
    }
    \caption{Two types of typical SONAR images. A polar image of area (a) is mapped to a corresponding area of the encoded polar image (b). The encoded image includes range($r$) and azimuth($\theta$).
    \label{fig:SONAR_geometry}
    }
\end{figure}

\begin{figure*}[t!]
	\centering
	\def\width{0.85\textwidth}%
    {%
		\includegraphics[clip, trim= 0 80 0 75, width=\width]{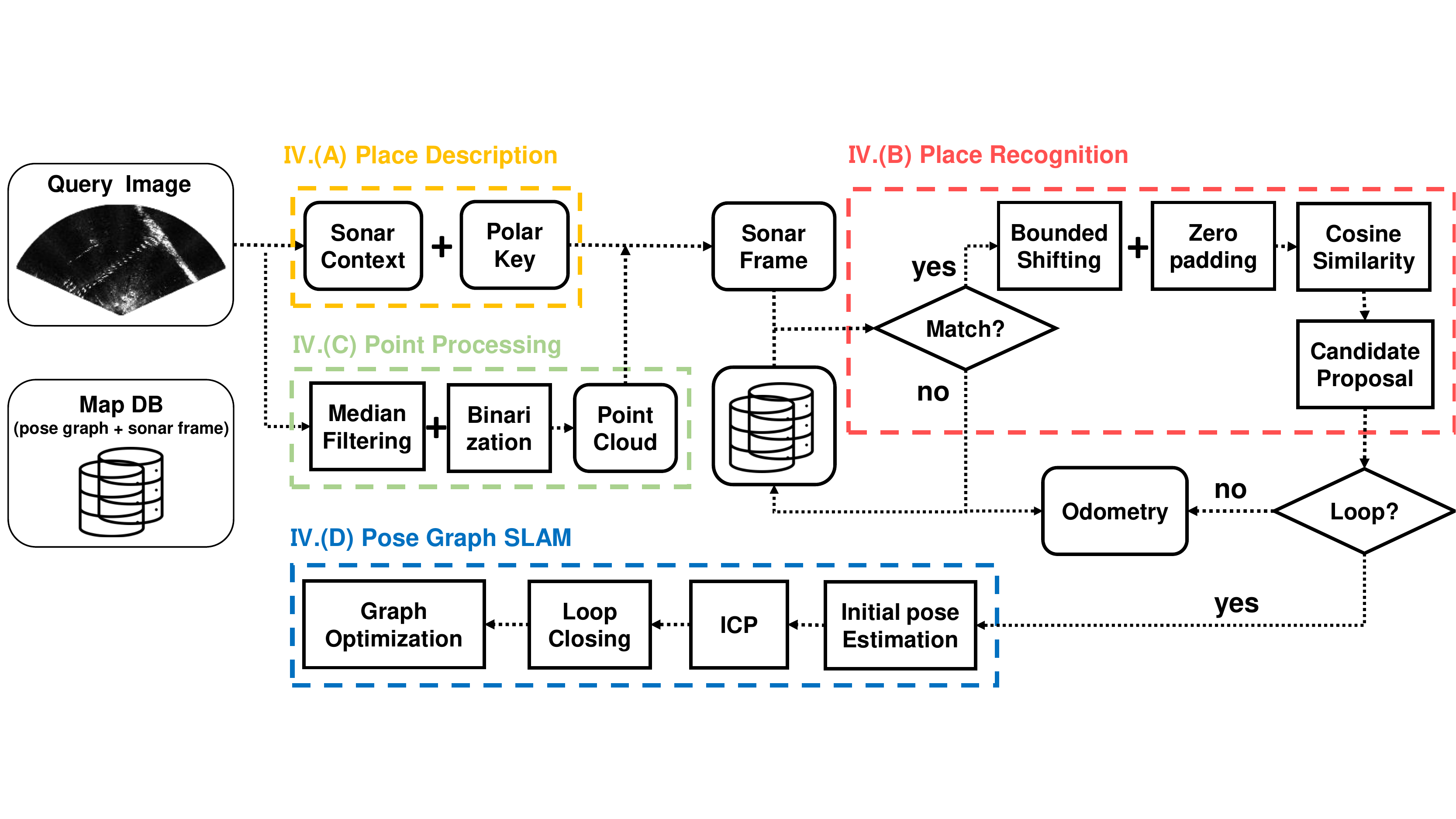}
	}
	\caption{Our proposed loop closure detection pipeline. Place description and point cloud processing are conducted in parallel. The place description part defines the SONAR context and polar key. 
	Place recognition finds the candidate via polar key, applies adaptive shifting, and compares cosine similarity between the query and candidate. Finally, loop closing is achieved using ICP.
	}
    \vspace{-0.5cm}
	\label{fig:our_slam_figure}
\end{figure*}

A single SONAR measurement $p_s$ is defined by \equref{equ:SONAR_measurements}, where $x_s, y_s$, and $z_s$ represent a point of SONAR measurement and $r,\theta$, and $\phi$ refer to the range, azimuth, and elevation in spherical coordinates, respectively.
\begin{equation}
	\label{equ:SONAR_measurements}  
		p_s= \begin{bmatrix}
                           x_s\\
                           y_s\\
                           z_s
                           \end{bmatrix}
                         = \begin{bmatrix} 
                           r\cos{\theta}\cos{\phi}\\
                           r\sin{\theta}\cos{\phi}\\
                           r\sin{\phi}
                           \end{bmatrix}
\end{equation}

However, elevation loss ($\phi=0$) occurs in the SONAR image. Therefore, inevitably, we obtain the SONAR measurements of polar coordinates as described in \equref{equ:SONAR_image} and \figref{fig:SONAR_geometry}
\begin{equation}
	\label{equ:SONAR_image}
	\mathcal{\hat{I}}(p_s)= \begin{bmatrix}
                           u\\
                           v
                           \end{bmatrix}
                        =  \begin{bmatrix}
                           \alpha\cdot \arctan{\frac{y_s}{x_s}}\\
                           \beta\cdot \sqrt{x^2_s + y^2_s}\\
                           \end{bmatrix}
\end{equation}
where $\alpha, \beta$ is the scale factor.

Then, the encoded polar image consists of $\mathcal{W} \times \mathcal{H}$ as described in \equref{equ:SONAR_space}

\begin{equation}
	\label{equ:SONAR_space}
   \mathcal{\hat{I}}(p_s) \in\mathbb{R}^{\mathcal{W} \times \mathcal{H}}
\end{equation}
where $\mathcal{W}$ and $\mathcal{H}$ are the image's width and height, respectively.

\section{Proposed Method}
In this section, we describe the details of the proposed method. The overall flow chart of the method is illustrated in \figref{fig:our_slam_figure}. In the figure, angled and rounded rectangles represent algorithms and data.

\subsection{Place Description: SONAR Context and Polar Key}

Based on the sensor properties, we propose a SONAR context for underwater environments, inspired by the scan context \cite{kim2021scan}. It divides the \ac{LiDAR} region into range and azimuth and designates the point on the highest z-axis as the representative of each region to summarize the surrounding structures in urban environments.
In underwater, the intensity of the SONAR image is a signal magnitude reflected from an object. Therefore, high intensity implies the inclusion of structural information, and we propose a SONAR context for a global descriptor utilizing the SONAR image as it is, encoding it into range and azimuth, and selecting the highest intensity as representative of each region.

To define the SONAR context, we first determine a single patch, $\mathcal{P}_{ij}$, which splits the SONAR image and consists of the patch size $p_w \times p_h$ as in \equref{equ:patch}. In this paper, we set $p_w = 4$ and $p_h = 4$, respectively.
\begin{equation}
	\label{equ:patch}
    \mathcal{P}_{ij} \in\mathbb{R}^{{p_w} \times {p_h}}
\end{equation}

Then, we find the highest intensity in each patch $\mathcal{P}_{ij}$ for the representative value of the patch, as in \equref{equ:max_intensity}.
\begin{equation}
	\label{equ:max_intensity}
    \mathcal{M}(\mathcal{P}_{ij}) = \max\limits_{(p \in \mathcal{P}_{ij})} i(p)
\end{equation}

Once the $\mathcal{P}_{ij}$ is decided, SONAR context $\mathcal{I}$ occupies the $A\times R$ space

\begin{equation}
	\label{equ:socon}
    \mathcal{I} \in\mathbb{R}^{{A} \times {R}}
\end{equation}
where $A$, and $R$ represent ${\mathcal{W} \over p_w}$, and ${\mathcal{H} \over p_h}$, respectively.

Finally, the SONAR context is defined by the following, as represented in \equref{equ:soconpatch}.
\begin{equation}
	\label{equ:soconpatch}
    \mathcal{I} = \bigcup_{i \in A, j \in R} \mathcal{\chi}_{ij}, \quad \mathcal{\chi}_{ij} = \mathcal{M}(\mathcal{P}_{ij})
\end{equation}

To recognize the revisited place with SONAR contexts, we need to grasp the similarity between the SONAR contexts.
Even if the SONAR context has implied information, comparing all of the contexts one by one increases the computational burden. Therefore, to resolve the issues, we propose a 1-D vector representing each SONAR context called the polar key. To generate the polar key, first of all, we average the intensity values of each row ($\mathfrak{p_1}, ...,\mathfrak{p}_{A}$) \equref{equ:row}
\begin{equation}
	\label{equ:row}
	\mathfrak{P}_j = \mathcal{F}(\mathfrak{p}_{1j}, ...,\mathfrak{p}_{Aj}) \qquad  j=1,...,R \\
\end{equation}
where the $\mathcal{F}(\cdot, \cdot)$ metric represents the average function.

Finally, by listing each value, we create a polar key \equref{equ:polarkey} composed of $R$ dimension vectors.

\begin{equation}
	\label{equ:polarkey}
	\mathfrak{P} = (\mathfrak{P}_1, ... , \mathfrak{P}_R)
\end{equation}

\subsection{Place Recognition}
Instead of comparing all of the contexts one by one, we can implement a light and fast searching algorithm by using the polar key, which is a vector that contains high intensity with structural characteristics.
By comparing the Euclidean distance between the polar key of a query node (current frame) and all polar keys of candidate nodes (previous frames excluded recently visited nodes), we can construct a KD-tree, used in the loop candidate proposal.
Thus, the first node of the KD-tree is the closest node, and then we determine it as the loop candidate. After the polar key specifies the loop candidate, the similarity method between a query context $\mathcal{I}^q$ and the specified loop candidate context $\mathcal{I}^c$ is adapted to determine the exact loop. 

To make this determination, we utilize the column-wise cosine distance method. This method entails dividing the query and candidate into columns ($c^q_j$, $c^c_j$ is $j$th column of $\mathcal{I}^q$ and $\mathcal{I}^c$) and determining the mean of cosine distance between columns in the same index for the query and candidate.

\begin{equation}
	\label{equ:cossim}
	\mathcal{D}_a\mathcal{(I}^q,\mathcal{I}^c) = {1 \over A} \sum\limits_{j=1}^{R} \left(1 - {c^q_j\cdot c^c_j \over ||c^q_j||\cdot ||c^c_j||}\right)
\end{equation}

In underwater environments, the \ac{AUV} is free to rotate. Thus, there is a high probability of seeing at a different angle and a resulting discrepancy in the distance, and it is not recognized as the revisited place, even though it remains in the same place. Therefore, we design augmented descriptions with a bounded shifting algorithm to achieve robustness in rotational and translational motion. Also, we bound the shifting range to prevent matching through less information.

We describe our adaptive shifting method in detail below.
\begin{itemize}
    \item To achieve robustness in rotational change, we conduct bounded-column shifting by setting a range suitable for the characteristics of the SONAR sensor and shifting the column in the row direction as described in \equref{equ:bcs}, where $\mu$ $(0 < \mu \leq 1)$ is the bounded-column factor.
    \begin{equation}
	    \label{equ:bcs}
    \mathcal{S}_a(\mathcal{I}^q,\mathcal{I}^c)= \min\limits_{n \in [-\frac{A}{2}, \frac{A}{2}]} \mathcal{D}_a\mathcal{(I}^q,\mathcal{I}^c_{n\times \mu})
    \end{equation}
    \item To supplement column shifting, we can also accomplish robustness in translation through bounded-row shifting that shifts the row in a column direction. This method entails dividing the query and candidate into rows ($r^q_i$, $r^c_i$ is $i$th row of $\mathcal{I}^q$ and $\mathcal{I}^c$) and determining the mean of cosine distance between columns in the same index as described in \equref{equ:row_cossim} and \equref{equ:brs}, where $\omega$ $(0 < \omega \leq 1)$ is the bounded-row factor.   

    \begin{equation}
	\label{equ:row_cossim}
	\mathcal{D}_r\mathcal{(I}^q,\mathcal{I}^c) = {1 \over R} \sum\limits_{i=1}^{A} \left(1 - {r^q_i\cdot r^c_i \over ||r^q_i||\cdot ||r^c_i||}\right)
\end{equation}

    \begin{equation}
	    \label{equ:brs}
    \mathcal{S}_r(\mathcal{I}^q,\mathcal{I}^c)= \min\limits_{m \in [-\frac{R}{2}, \frac{R}{2}]}
    \mathcal{D}_r\mathcal{(I}^q,\mathcal{I}^c_{m\times \omega})
    \end{equation}

    Encoded polar images include range and azimuth information, explicitly. Therefore, if there is a significant difference in the SONAR image's column direction, the difference in distortion will increase, and it cannot exist at the same distance. Therefore, our method sets a lower bounded-row factor $\omega$ and captures translation change.

    \item Considering \ac{SONAR}'s \ac{FOV}, we apply zero padding on the shifted columns and rows ($c^c_0 = \bold{0}$ and $r^c_0 = \bold{0}$) to prevent circular shifting to the opposite side. When rows and columns shift in the same sign direction(i.e. left and right, respectively), the rows and columns that shifted persist to zero.
    
\end{itemize}

As a result, we can find the revisited place with the most similar SONAR context while shifting each column and row and comparing cosine similarity to the specified threshold. Also, the shifted column and row steps will be utilized for the initial pose in loop closure.

\subsection{Point Processing}
In parallel with the SONAR context description, we also retrieve point clouds from SONAR images. Due to the speckle noise in the SONAR image, we apply several image enhancement methods. To select a reliable point, we first apply a median filter to the image in order to be robust to the noise further. Then, Otsu's binarization \cite{otsu1979threshold} is utilized to extract point clouds. In practice, Otsu's binarization partially compensates for the detail loss from median filtering by specifying that the distribution of light and shade is the most uniform. Finally, we select the set of points ($\mathcal{C}$) in the medium section in the column direction and construct the SONAR frame $\textit{S} = (\mathfrak{P}, \mathcal{I}, \mathcal{C})$ for the overall \ac{SLAM} pipeline.

\subsection{Loop factors for Pose-graph \ac{SLAM}}
Now, we perform a \ac{SLAM} algorithm using the SONAR frame. Our proposed method is based on the pose graph \ac{SLAM} function that minimizes the drift error as below.
\begin{equation}
    \label{equ:pose_estimation}
    \begin{aligned}
        X^* & = \argmin_{X}\sum_t\left\| f(x_{t}, x_{t+1}) - z_{t,t+1} \right\|_{\sum_{t}}^2 \\
        & \qquad \qquad+ \sum_{i,j\in LC}\left\|f(x_{i}, x_{j}) - z_{i,j} \right\|_{\sum_{i,j}}^2
    \end{aligned}
\end{equation}


In this function, the nodes $X = [x_1^T, \;\cdots, \;x_t^T]^T$ are the frame's 6-DOF poses ($x_t = [x, y, z, r, \theta, \phi]$) at time $t$, corresponding to SONAR frames. $f(\cdot, \cdot)$ metrics estimate the two sequential 6-DOF relative poses. Odometry-based relative 6-DOF pose constraints are defined as $z_{t,t+1}$. For loop closure ($z_{i,j}$), we construct a 3-DOF (XYH) constraint by applying 2-D \ac{ICP} scan matching. Also, we utilize the initial pose from the context matching process for robust estimation of the relative pose. 

\section{Experimental Results}

\subsection{Dataset and Experiment settings}
We use HOLOOCEAN \cite{potokar2022holoocean}, KRISO water tank \cite{jang2021multi}, and ARACATI 2017 datasets to show that our method can be adapted to various environments.

\begin{figure*}[!ht]
	\centering
	\def\width{1.95\columnwidth}%
	\includegraphics[clip, trim= 40 10 20 10, width=\width]{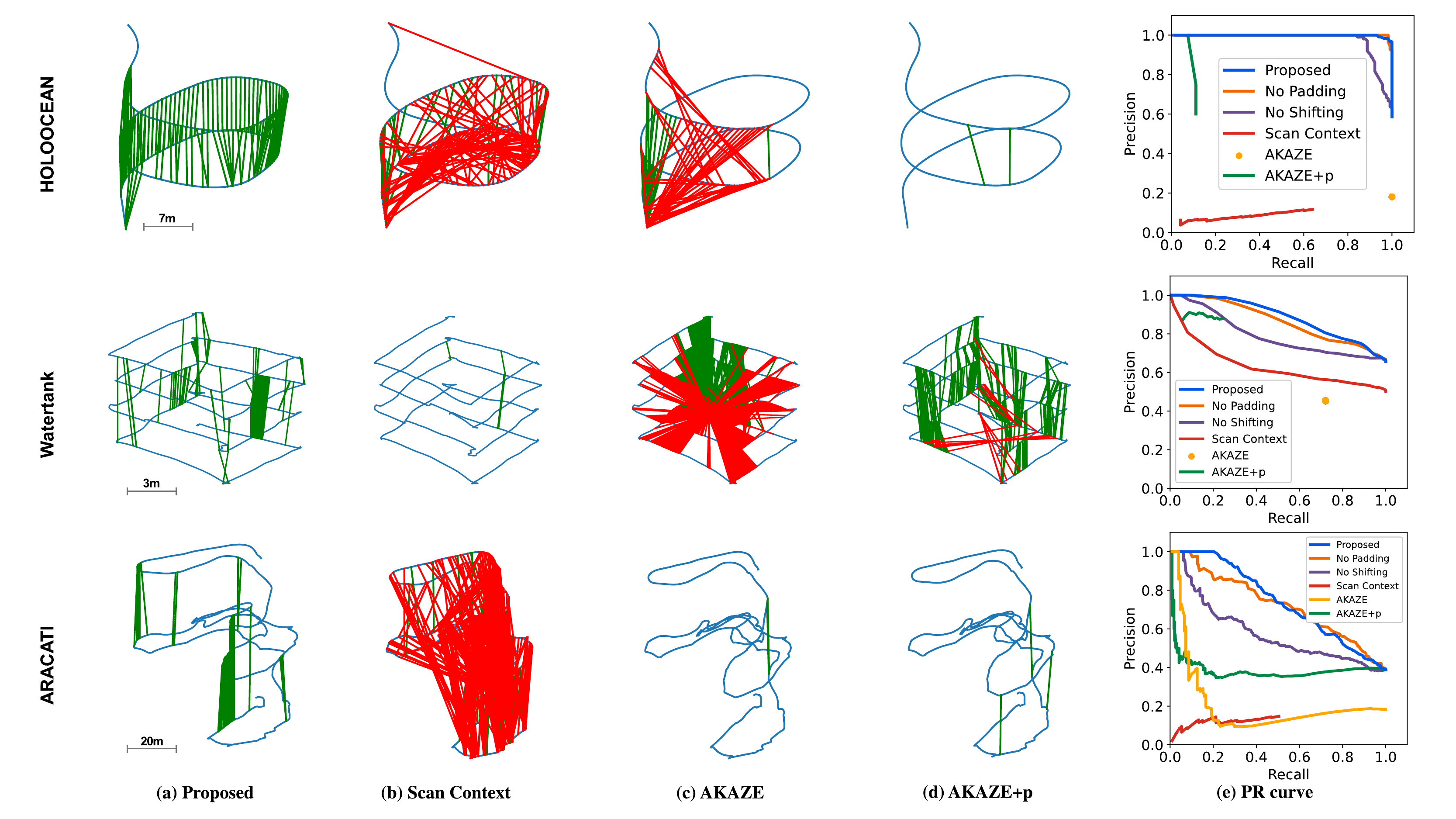}
	\caption{Time-elevation trajectory with correct(green) and incorrect(red) matching for various  methods and their precision-recall curves including self-ablated methods. We show these qualitative results when each method has the highest possible level of precision as much as possible. In detail, AKAZE+p is the method achieved by global image search with a polar key and the distance based on the AKAZE inlier, whereas AKAZE is the method achieved by one-by-one feature matching with no polar key involved.
	}
        \vspace{-0.2cm}
	\label{fig:total}
\end{figure*}

\subsubsection{HOLOOCEAN \cite{potokar2022holoocean}} Among various simulated environments, we select OpenWater with a size of 2km containing a sunken submarine and many rolling hills. The dataset is obtained by traveling in a circular route twice. There is a slight difference in angle and translation between the two routes.

\subsubsection{KRISO watertank \cite{jang2021multi} } $7m \times 7m$ square route dataset obtained by \ac{DIDSON} in the real water tank with five artificial markers. As we conducted the experiment in the water tank, the range is limited to the tank's bottom. Because a  KRISO water tank does not include the ground-truth trajectory, we use the trajectory \cite{jang2021multi} utilized as the ground truth.

\subsubsection{ARACATI 2017 \cite{dos2022cross}} Data were collected on the marina of the Yacht Club of Rio Grande - Brazil using a remote-operated vehicle LBV 300-5 from Seabotix on an unfixed route. The ARACATI dataset uses the Blue View P900-130, which has a range of 50m and a depth of $130^{\circ}$. GPS measurement is regarded as a reference because the data were collected by holding an underwater vehicle on a floating boat. The marina has a depth of 1-5m, and the coast is covered with stone. 

We compare and evaluate our method against AKAZE method \cite{Alcantarilla2013FastED} most frequently used underwater \cite{li2018pose} \cite{jang2021multi}, AKAZE with polar key (AKAZE+p), which is an AKAZE-based method combined with our proposed polar key, and original scan context \cite{kim2018sc}. To evaluate the performance of AKAZE-based description, an inlier ratio of feature matching is utilized with brute-force searching for loop detection. AKAZE+p finds a matching pair with the polar key and checks the similarity given loop candidates. To evaluate the scan context, we first convert the image into point clouds of x, y, and intensity and find a matching pair. 
As an ablation study, we also validate the shifting module, essential for rotational robustness, and the padding module that considers \ac{SONAR}'s \ac{FOV}.
For all methods, we consider the true positive matching pair if the distance between the query and ground-truth pose is less than the predefined value. We choose the appropriate distance according to the scale of each dataset (Holoocean: 3m, Watertank: 2m, and Aracati: 6m). The proposed method is implemented in Python, and all experiments are carried out on the same system with an Intel i7-12700 KF at 3.60GHz and 32GB memory.

\subsection{Precision and recall evaluation}
In \figref{fig:total}, we first evaluate the proposed method against the previous method by visualizing the time elevation plots of true (green) and false (red) matches. All figures are plotted at the maximum precision for all methods. In the figure, our method successfully found loops for different types of environments. 

Also, in \figref{fig:total}(e), we evaluate our method against others using a precision-recall curve. First, AKAZE shows poor performance in all datasets, implying that using the feature-matching method is challenging in underwater environments. However, it is noteworthy that the polar key can enhance recognition ability, considering the better performance of AKAZE+p. In contrast to the feature method, our approach outperforms all other methods, including ablation results. Especially in the ARACATI dataset, our module shows outstanding performance for rotational and translational variance compared to any other method.

\subsection{Partial Overlap}
\begin{figure}[h]
	\centering
	\includegraphics[clip, trim=10 85 10 27, width=\columnwidth]{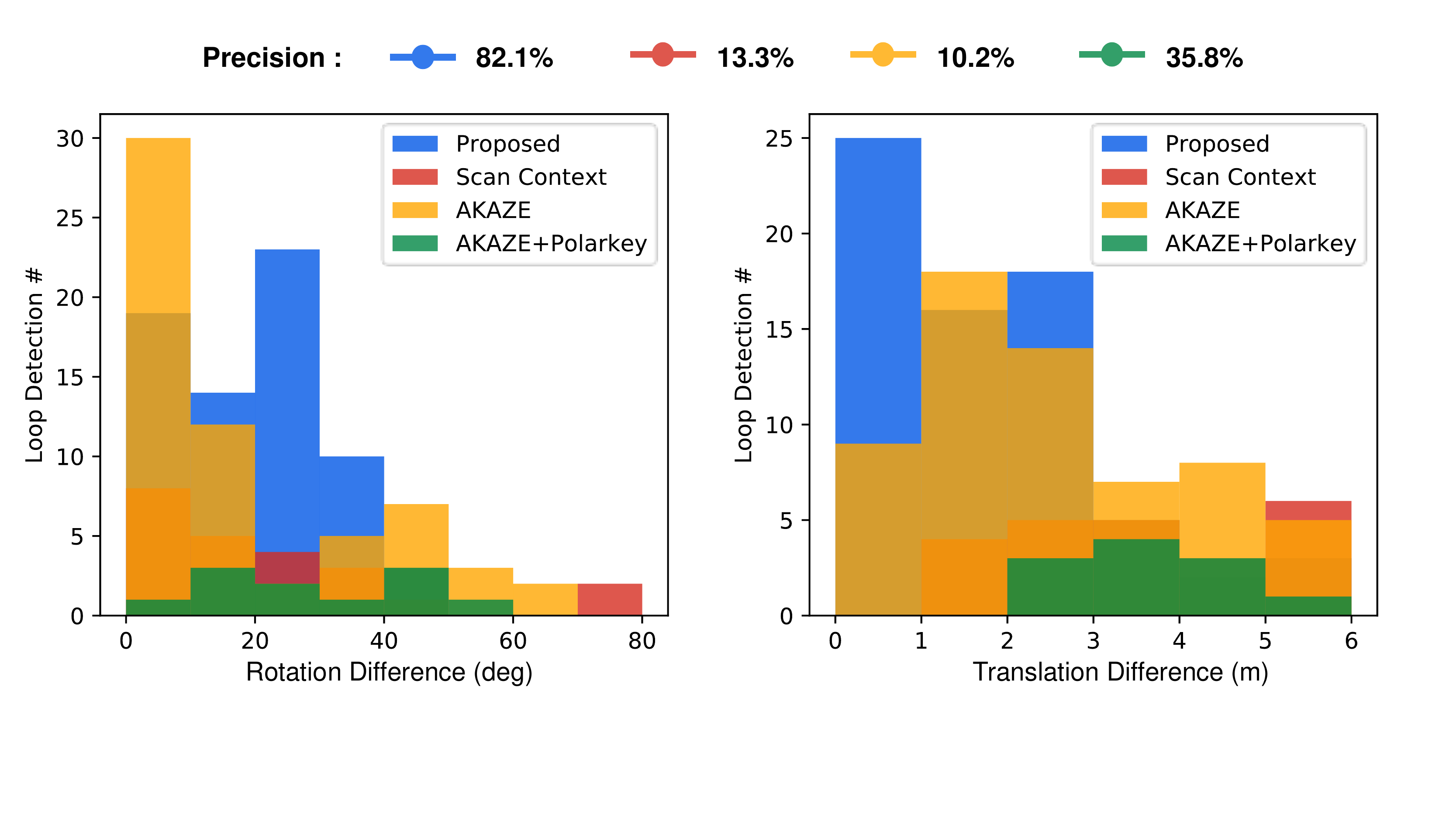}
	\caption{Histogram of detected loop pairs by rotation and translation difference. The figure is plotted at the recall is 0.4 for each method.
}
    \vspace{-0.3cm}
    \label{fig:partial_overlap}
\end{figure}

\figref{fig:partial_overlap} shows a numerical histogram of detected loop pairs (between the query and candidate) by rotation and translation difference in the ARACATI dataset.

\begin{figure*}[t]
    \centering
    \def\width{0.6\columnwidth}%
    \subfigure[HOLOOCEAN]{
    \includegraphics[width=\width]{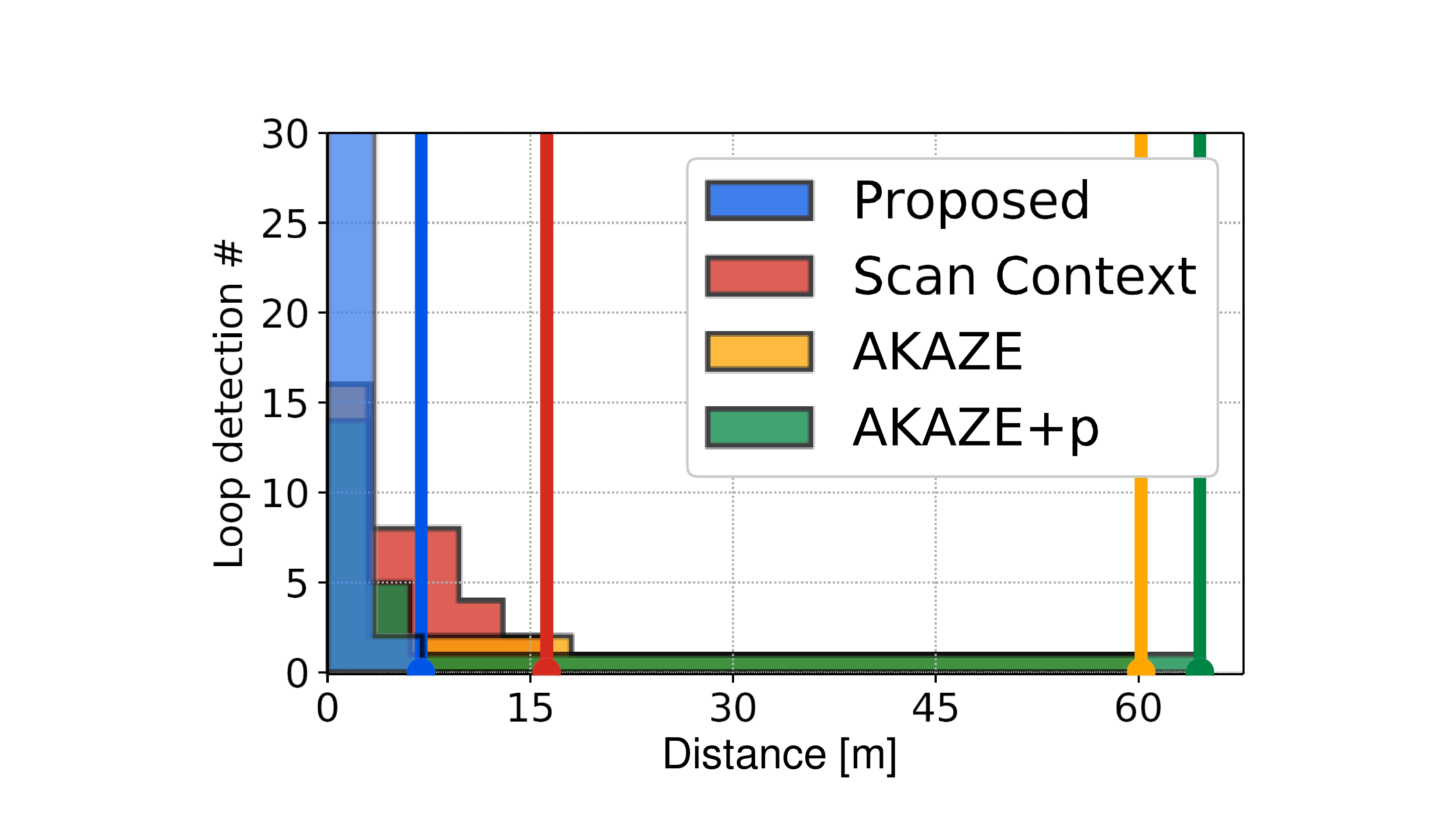}
    \label{fig:loop_gap:holo}

    }
    \subfigure[KRISO]{
    \includegraphics[width=\width]{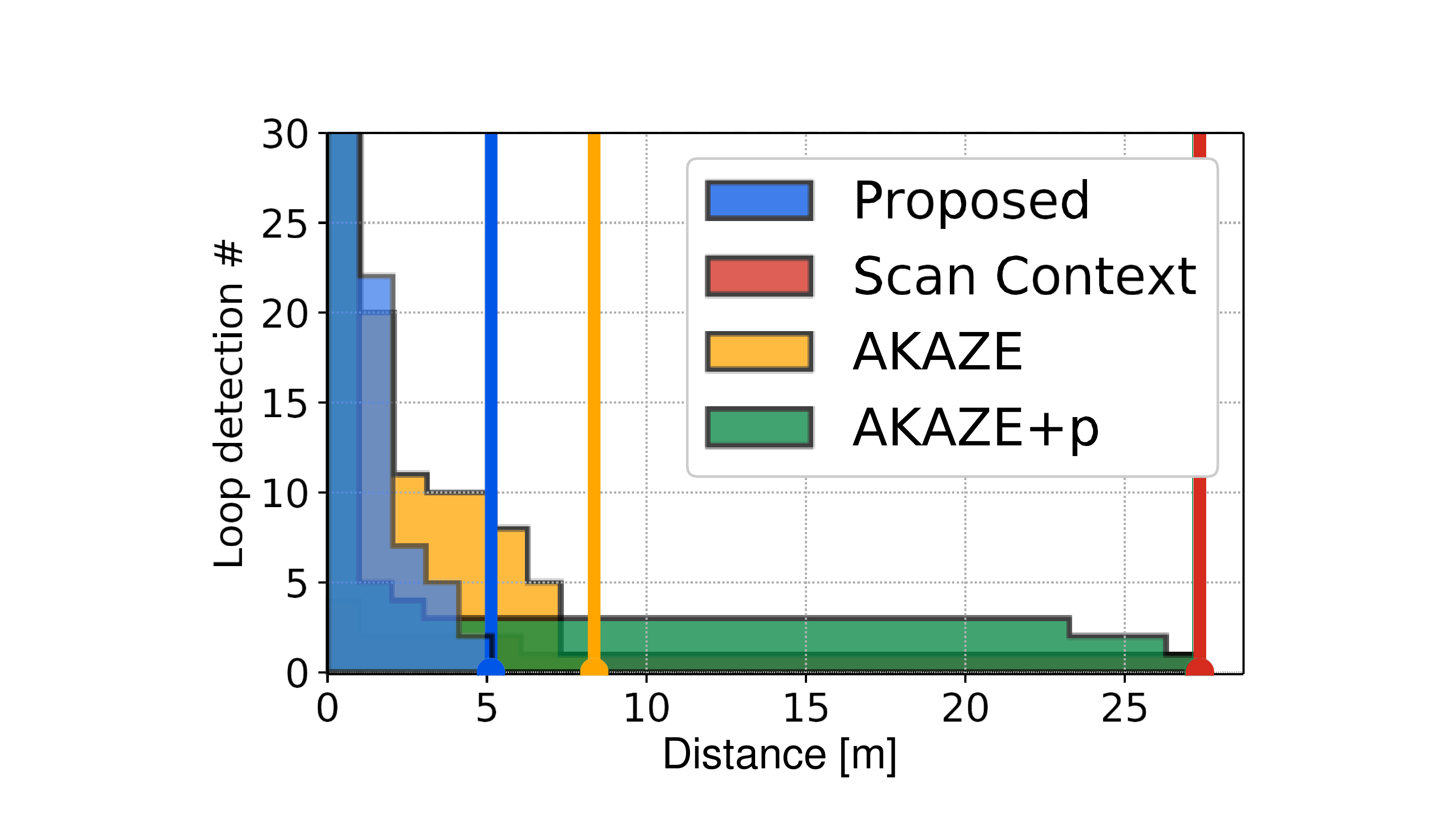}
    \label{fig:loop_gap:kri}

    }
    \subfigure[ARACATI]{
    \includegraphics[width=\width]{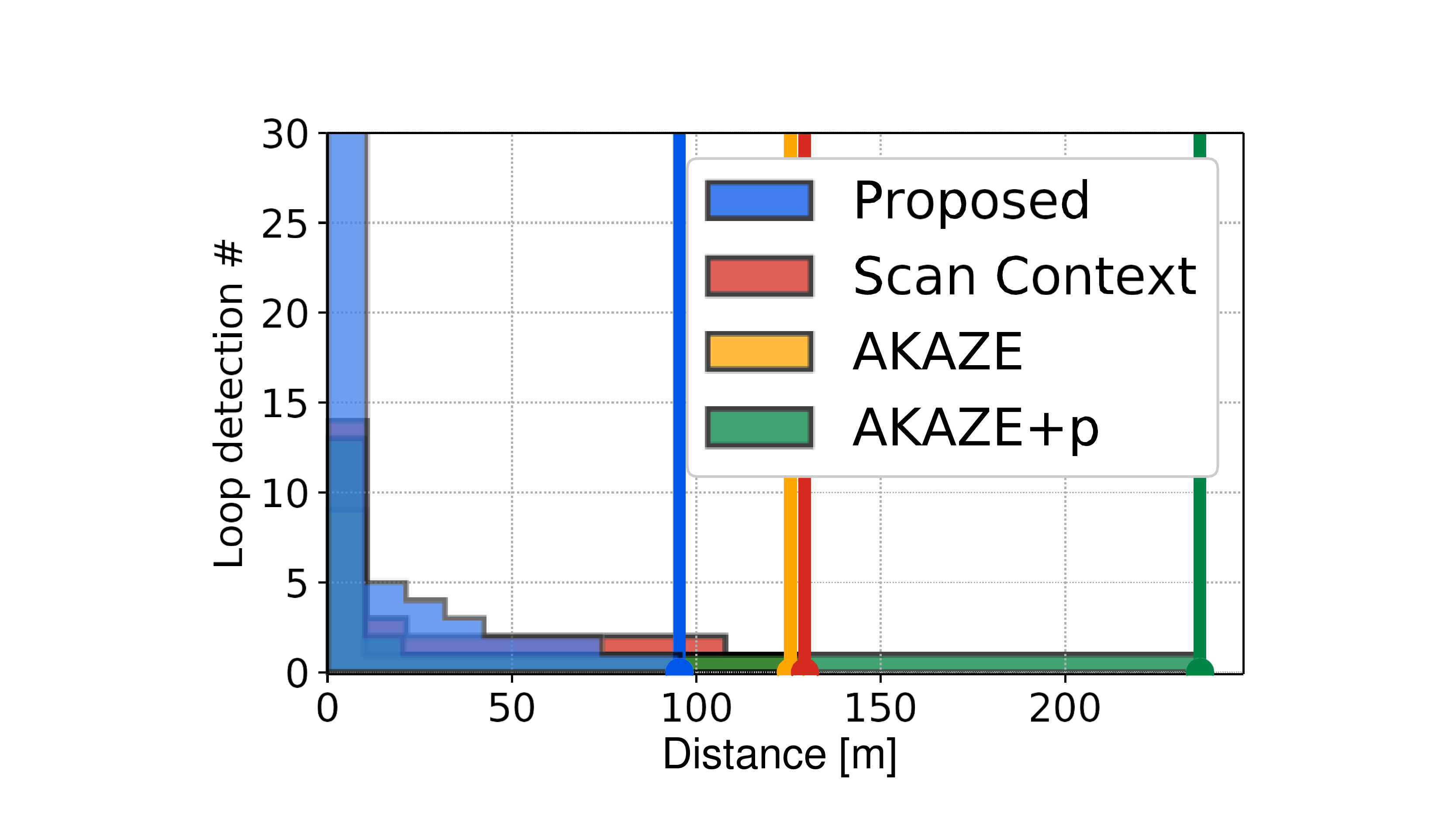}
    \label{fig:loop_gap:ara}

    }
    \caption{Distribution of the distance explored until the next loop detection occurs. Our method preserves continuous loop closures over the traverse with minimum blind traversal. The straight lines represent the maximum distance without loop closures (lower is better).
    }
    \label{fig:loop_gap}
    \hspace{-4mm}
    \vspace{-0.3cm}

\end{figure*}

The SONAR context utilizes adaptive shifting to estimate relative pose, showing generally robust performance on rotation and translation in the number of detected loops and variance.
The proposed method can capture about $40^{\circ}$ rotation and 5m translation differences with 80 $\%$ precision. Also, we think it is a suitable performance because of the applied bounded shifting on the context matching procedure. Although there are some positive matchings of other methods from $60^{\circ}$ to $80^{\circ}$ on rotation differences, the precision of our method is significantly higher than other methods. 

\subsection{Blind Traversal}
If \ac{AUV} traverses a longer distance without loop closure, the uncertainty and error of the robot state significantly increase. \figref{fig:loop_gap} shows the distribution of traveling distance without loop closure named Blind Traversal by accumulating the distance between consecutive true positive matching. Therefore, the narrow distribution form near the origin and lower maximum distance represent better performance. The figure shows that the proposed method results in continuous and abundant loop closures. This represents that the proposed method preserves reliable and robust loop closures during robot navigation.

\subsection{Global Pose Estimation with Loop Closures}
\begin{figure}[h]
	\centering
	\def\width{0.46\columnwidth}%

	\subfigure[Without initial pose]{
    \includegraphics[clip, trim= 15 0 45 10,width=\width]{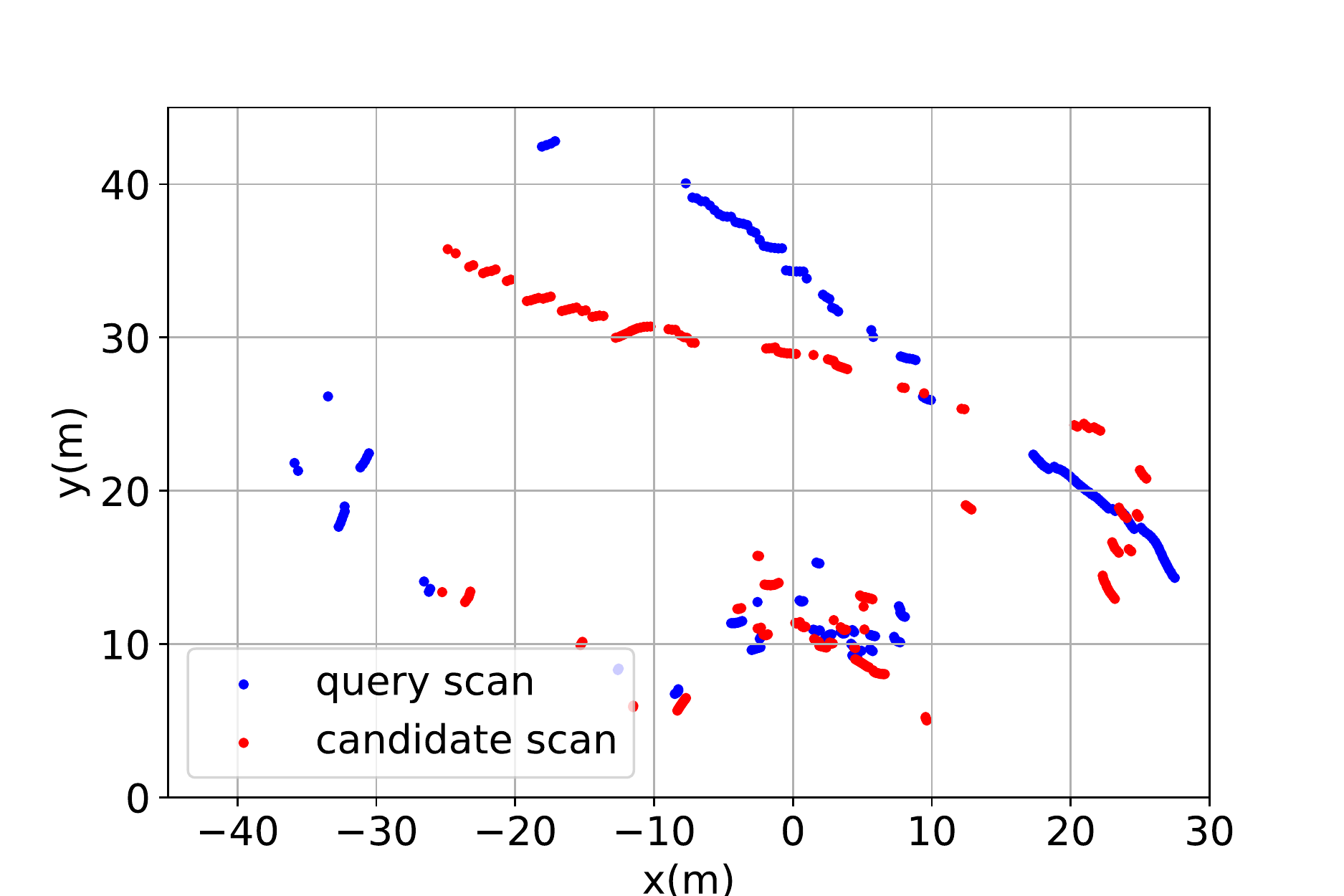}
    \label{fig:icp1}

    }
    \subfigure[With initial pose]{
    \includegraphics[clip, trim= 10 0 45 10,width=\width]{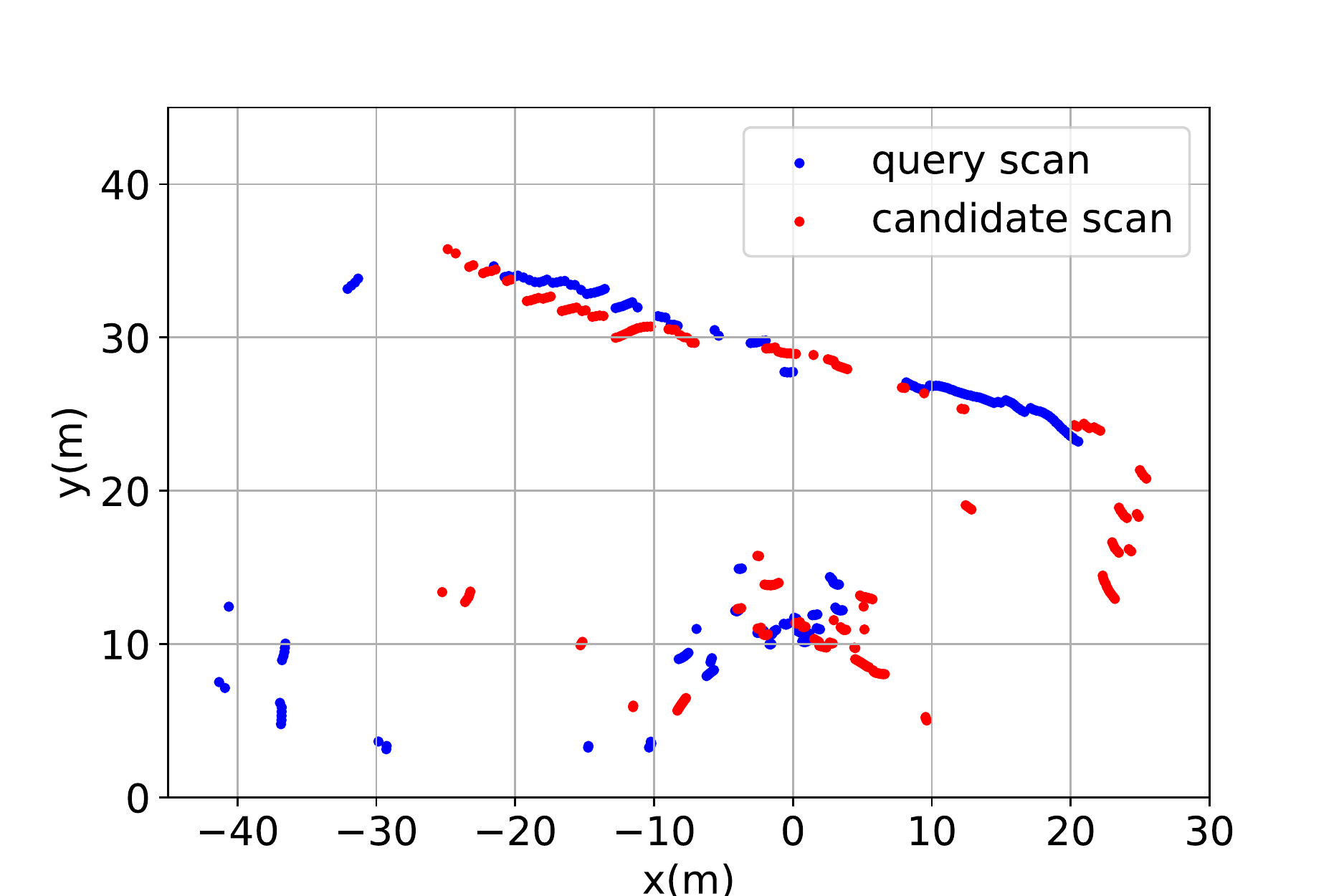}
    \label{fig:icp2}
    }

	\caption{Applied relative pose estimated from adaptive shifting in \ac{ICP}. The figure on the left is plotted point cloud without the initial pose, and the figure on the right is after the initial pose is applied.
	}
	\label{fig:icp_points}
\end{figure}

To prove the applicability of the SONAR context in real underwater environments, we evaluate the effectiveness of the proposed method to the entire SLAM pipeline with the ARACATI dataset. Given odometry measurements, we verify the loop closure factors against the reference poses. To extract accurate relative motion between query and retrieved SONAR frames, we apply the initial pose from the context matching process. \figref{fig:icp_points} shows the guidance of the initial pose update before point cloud registration. 
Because the result of \ac{ICP} is often disturbed by poor initialization,  pre-aligning point clouds leads to better registration performance. 

\begin{figure}[h]
    \centering
    \vspace{-0.1cm}
    \subfigure[Trajectories]{
        \includegraphics[clip, trim=110 295 120 310, width=0.65\columnwidth]{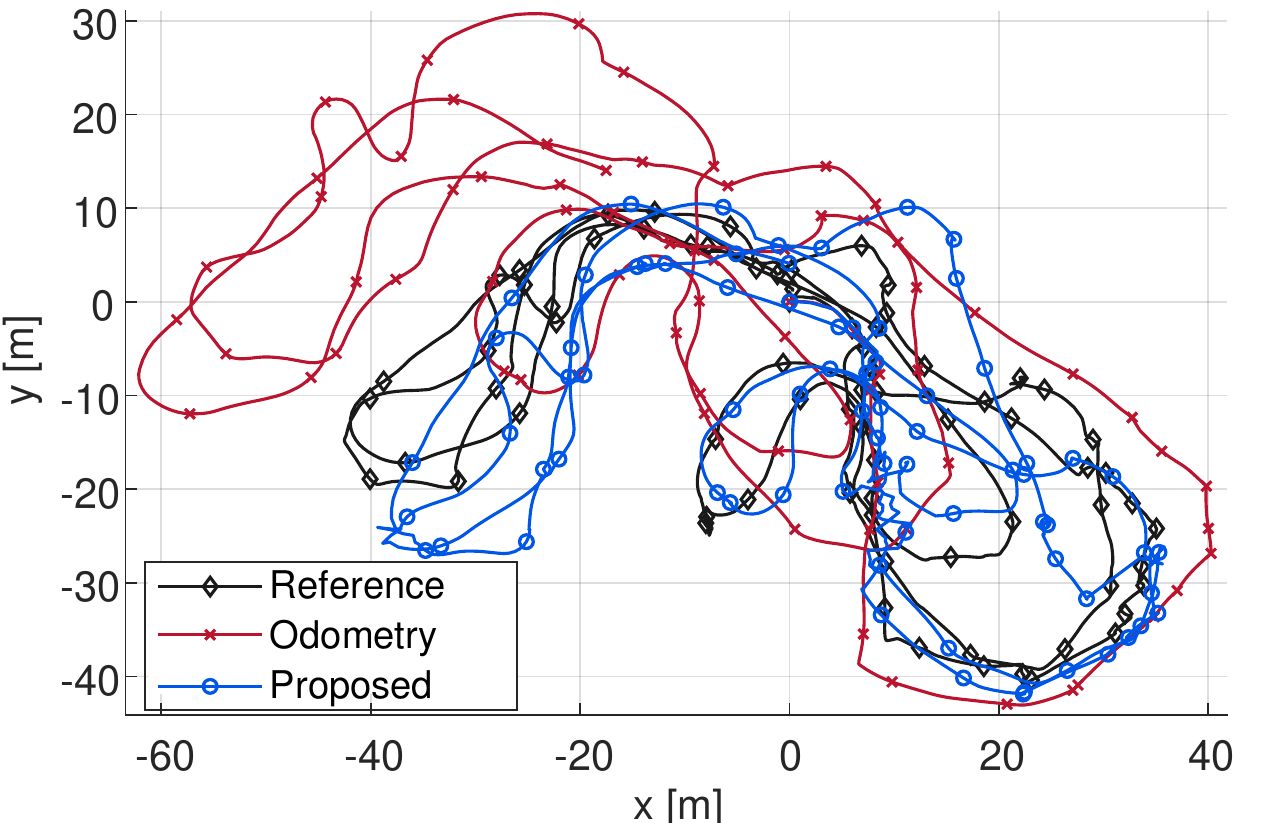}
        \label{fig:trajectory}
    } \\
    \subfigure[Error Plot]{
        \includegraphics[clip, trim=130 310 130 320, width=0.65\columnwidth]{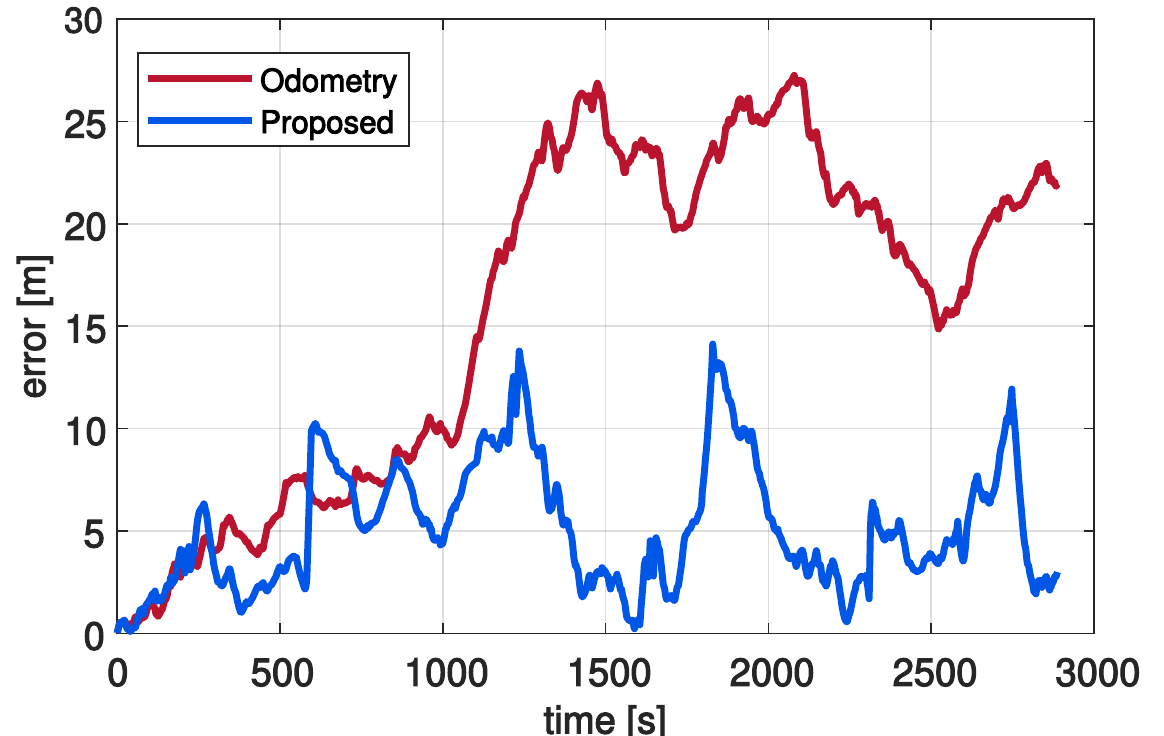}
        \label{fig:loc_error}
    }
	\caption{Global robot trajectory and error plot. (a) A bird-eye view of three trajectories. (b) Localization errors corresponding to the reference pose. Compared to the odometry error, we can observe the localization error of our method dropped by loop closure.}
    \label{fig:icp_trajectory}
    \vspace{-0.2cm}
\end{figure}

In \figref{fig:icp_trajectory}, we leverage the detected loop closure and estimated relative pose using the SONAR context and correct the accumulated drift error. 
The SONAR context detects more loops, especially in coastlines or offshore structures, and has rotation-robustness characteristics by estimating relative angular differences precisely. 
To compare the performance of underwater global localization, we refer to \cite{dos2022cross} which has the trajectory evaluation result of the ARACATI dataset. Compared to the error chart in \cite{dos2022cross}, we  find that the proposed method preserves stable and accurate localization results for the sequence.

\section{conclusion}

We propose an imaging SONAR-based global descriptor, the SONAR context, for robust place recognition in various underwater environments. The proposed method explicitly describes the geometric characteristics of surrounding environments. We design adaptive shifting and matching procedures by considering SONAR characteristics and propose further utilization of the context with loop closure factors. Compared to existing approaches, our method shows outperforming results across various datasets and metrics. 

In future work, we plan to modify the SONAR context of different types of SONAR sensors, such as side-scan SONAR and profiling SONAR. By developing hierarchical representation or embedding semantic information, we will extend the SONAR context for multi-session SLAM and map management. 



%


\bibliographystyle{packages/IEEEtranN} 
\bibliography{packages/string-short, packages/reference}

\end{document}